\documentclass[runningheads]{llncs}
\usepackage[T1]{fontenc}
\usepackage{graphicx}
\usepackage{subfigure}
\usepackage{booktabs}
\usepackage{makecell}
\usepackage{tabularx}
\usepackage{multicol}
\usepackage{multirow}
\usepackage{amsmath}
\usepackage{xcolor}
\usepackage{amssymb}
\usepackage {amsmath}
\usepackage{bbding}
\usepackage{url}
\usepackage{diagbox}
\usepackage{graphicx,verbatim}
\begin{document}
%
\title{Learning to Distort: Weakly-Supervised Image Quality Transfer for Prostate DWI Correction}
\titlerunning{Weakly-Supervised Image Quality Transfer}
%

\author{
Yucheng Tang\inst{1} \and
Wen Yan\inst{1} \and
Alexander Ng\inst{2} \and
Natasha Thorley\inst{3} \and
Pawel Rajwa\inst{2} \and
Yipei Wang\inst{1} \and
Aqua Asif\inst{4,2} \and
Clare Allen\inst{5} \and
Louise Dickinson\inst{5} \and
Francesco Giganti\inst{2,5} \and
David Atkinson\inst{3} \and
Shonit Punwani\inst{5,3} \and
Daniel Alexander\inst{6,7} \and
Shaheer Ullah Saeed\inst{1} \and
Veeru Kasivisvanathan\inst{2,8} \and
Yipeng Hu\inst{1}
}

\authorrunning{Tang et al.}

\institute{
UCL Hawkes Institute and Department of Medical Physics and Biomedical Engineering,
University College London, UK \\
\and
Division of Surgery and Interventional Science,
University College London, UK \\
\and
Centre for Medical Imaging,
University College London, UK \\
\and
British Urology Researchers in Surgical Training (BURST), UK \\
\and
Department of Radiology,
University College London Hospitals NHS Foundation Trust, UK \\
\and
Centre for Medical Image Computing,
University College London, UK \\
\and
Department of Computer Science,
University College London, UK \\
\and
Department of Urology,
University College London Hospitals NHS Foundation Trust, UK \\
}

\maketitle              
\begin{abstract}
Single-shot echo-planar prostate diffusion-weighted imaging (DWI) is frequently complicated by geometric distortions, which impact the ability to derive reliable diagnoses from such images. Developing automated correction methods is challenged by the absence of paired distorted and undistorted clinical scans. In this paper, we first propose a novel weakly-supervised image quality transfer (IQT) framework from undistorted to distorted images that utilizes image quality assessment (IQA) signals to supervise the transfer process. Unlike traditional methods that require expensive, voxel-wise paired data or resort to developing unpaired algorithms, our approach utilizes image-level quality labels (here, distorted vs. undistorted) to establish latent quality prototypes within a pre-trained feature space. 
Recognizing that simulating realistic distortions is more reliable than direct unpaired correction,
we describe a weakly-supervised prototype flow matching algorithm to explicitly regularize generative trajectories towards distorted prototypes, producing realistic susceptibility artifacts that mimic clinical degradations.
By synthesizing these realistic pairs, we enable a second IQT model to be trained in the forward direction for distortion correction.
Experimental results demonstrate that our generated images successfully mimic the diagnostic interference of real-world artifacts, which leads to more capable distortion correction IQT models. In addition to qualitative comparisons, we also conduct exhaustive quantitative evaluations that compare our approach with existing unpaired approaches (e.g., CycleGAN, UNIT-DDPM, and OT-FM) - as either forward or reverse alternatives - by assessing clinical downstream task performance in PI-RADS and Gleason score classification, using both in-distribution and external data sets. Code is available at \url{https://anonymous.4open.science/r/ProtoFM-IQT-06DD}.

\keywords{Prostate DWI  \and Flow Matching \and Image Transfer.}

\end{abstract}
\section{Introduction}
Prostate diffusion-weighted imaging (DWI) is playing a growing role in cancer detection, grading, risk assessment and their subsequent biopsy and treatment planning. However, single-shot echo-planar imaging is inherently prone to susceptibility artifacts caused by rectal air and patient motion~\cite{donato2014geometric}. These artifacts lead to geometric distortions, thereby misleading the spatial correspondence between DWI and anatomical T2-weighted images (T2WI)~\cite{rakow2015prostate} and, when used alone, its true zonal and pathology localisation with respect to surrounding anatomy. In clinical practice, distorted DWI often needs repeated scans, extending examination time and increasing patient and healthcare burden. 

Although acquisition-stage techniques~\cite{lawrence2022reduced,chien2025deep} and reverse phase-encoding tools such as TopUp~\cite{andersson2003correct} alleviate distortions effectively, they require additional scans or paired images with opposite encoding directions, which are often unavailable in routine clinical workflows.
Fully-supervised image quality transfer (IQT) methods~\cite{liao2018referenceless,hu2020distortion,bian2023drdisco} rely on paired distorted-undistorted scans, which are generally infeasible.
Unpaired IQT methods, including CycleGAN~\cite{sandfort2019data} and diffusion-based models~\cite{zou2025cyclediff,sasaki2021unit} may respectively face challenges in unstable training~\cite{johnson2021generative} and high inference latency~\cite{song2020denoising}, with additional risk in anatomical hallucinations~\cite{tivnan2024hallucination}.

To address these challenges, we propose a novel weakly-supervised IQT method that overcomes paired data scarcity. Using weak quality-based supervision, simulating distortions is found more reliable than direct correction (speculatively, generative mistakes (e.g., hallucinations) during degradation are less harmful than hallucinating non-existent or removing true anatomical structures during correction). Therefore, we first generate realistic distortions from undistorted inputs based on flow matching (FM)~\cite{lipman2022flow}, yielding a synthetic paired dataset to train a second fully-supervised distortion correction model.

Our contributions include:
First, a novel weak-IQA-supervised prototype FM method to generate realistic distorted DWI from undistorted inputs. 
Second, an effective IQT algorithm reformulating the problem as realistic distortion generation followed by fully-supervised correction.
Third, using two clinical downstream tasks on two independent data sets, we first demonstrate that classifiers evaluated on our synthesized distorted images exhibit significant performance degradation in PI-RADS and Gleason scoring, indicating strong correlation between the synthesized artifacts and diagnostic decision-making. Furthermore, we show that the generated distorted-undistorted pairs can be successfully used to train a second IQT model that effectively corrects real-world DWI distortions.

\section{Method} 

\begin{figure*}[t] 
\label{main} \centering \includegraphics[width=\textwidth]{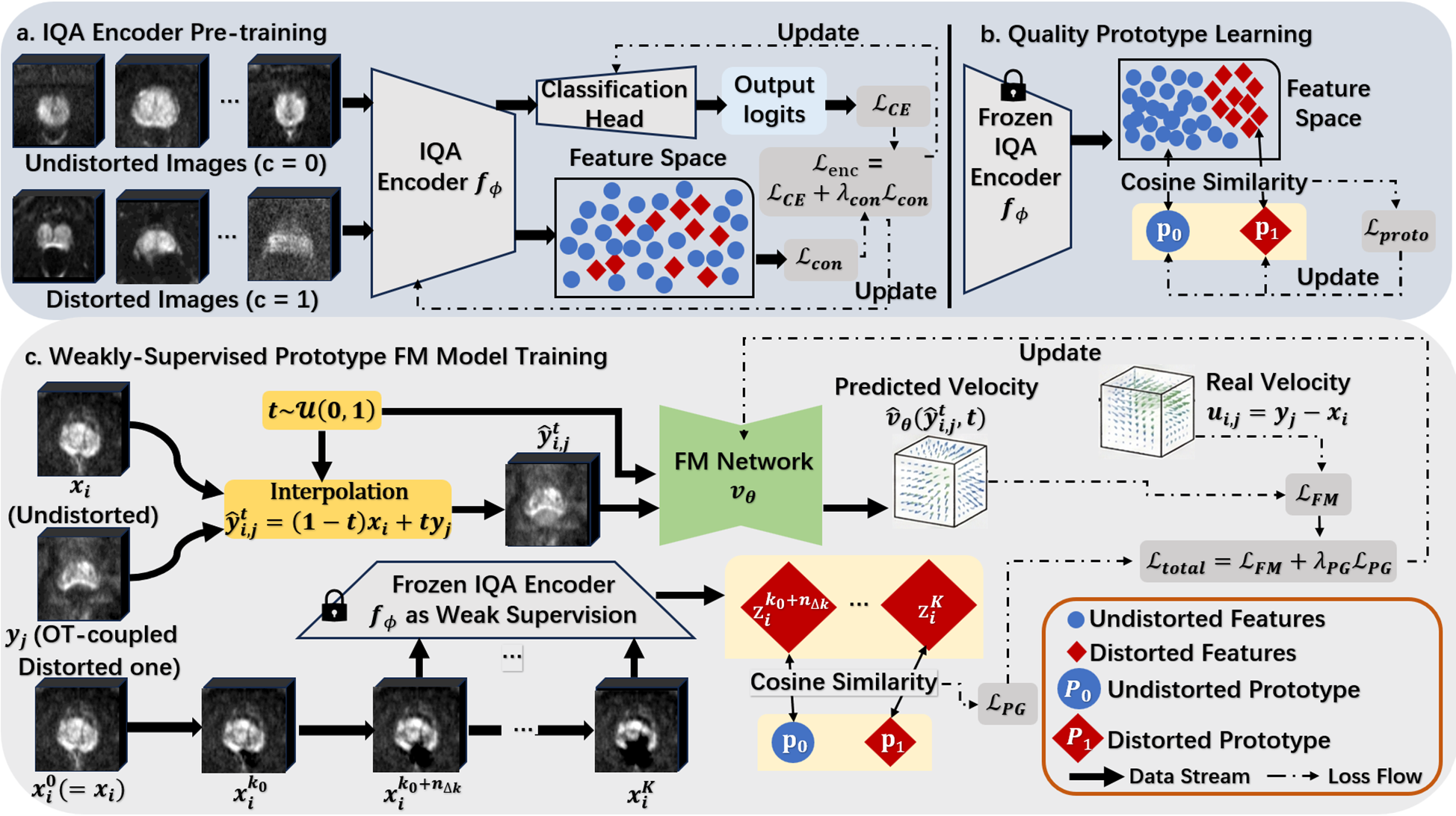} \caption{Overview of the proposed Weakly-Supervised Prototype FM framework. (a) An IQA encoder is pre-trained to construct a feature space that distinguishes distorted and undistorted images. (b) Quality prototypes are learned to represent the centers of these domain distributions. (c) Training of the FM-based IQT model, where the generation trajectory is explicitly guided toward the distorted manifold using the pre-trained encoder and learned prototypes.
} \end{figure*}

The proposed IQT method comprises three stages: 1) constructing a weak supervision signal via an IQA encoder and quality prototypes, as shown in Fig.~\ref{main}(a) and (b), and 2) training a FM network constrained by this signal, as shown in Fig.~\ref{main}(c). 3) Inference via the trained FM network and use the generated paired images to fully supervise a second IQT model.

\subsection{Construction of Weak Quality Supervision Signal}
\textbf{IQA Encoder Pre-training.}
We first pre-train an IQA encoder $f_{\phi}(\cdot)$, to discriminate distorted features from undistorted ones and extract distortion-related representations. The encoder accepts a 3D DWI volume $x_i$ as input and produces a global feature vector $\mathbf{z}_i \in \mathbb{R}^{d}$, where $d$ denotes the feature dimension.
Consider a mini-batch $\mathcal{B}$ of samples $\{ (x_i, c_i) \}_{i=1}^{|\mathcal{B}|}$, where $c_i \in \{0,1\}$ is the binary label with $c_i=0$ and $c_i=1$ representing undistorted and distorted images, respectively. The IQA encoder extracts the feature vector $\mathbf{z}_i$, which is then mapped to the output logits $\mathbf{o}_i$ via a linear classification head. The encoder and classification head are optimized using the standard binary cross-entropy loss $\mathcal{L}_{\mathrm{CE}}$ against the ground-truth labels $c_i$.
To encourage a well-structured feature space with compact intra-class clusters and clear inter-class separation, we incorporate a contrastive loss $\mathcal{L}_{\mathrm{con}}$. This loss maximizes feature similarity among samples of the same class while separating features from different classes. Given a temperature parameter $\tau$, the loss is formulated as in Eq. (\ref{con_loss})
\begin{equation}
\mathcal{L}_{\mathrm{con}} =
\frac{1}{|\mathcal{B}|}
\sum_{i=1}^{|\mathcal{B}|}
\frac{-1}{|\mathcal{P}(i)|}
\sum_{p \in \mathcal{P}(i)}
\log\frac{\exp(\cos(\mathbf{z}_i,\mathbf{z}_p)/\tau)}{\sum_{j \neq i}\exp(\cos(\mathbf{z}_i,\mathbf{z}_j)/\tau)
}
    \label{con_loss}
\end{equation}
where $\cos(\mathbf{a}, \mathbf{b}) = \frac{\mathbf{a} \cdot \mathbf{b}}{|\mathbf{a}| |\mathbf{b}|}$.
$\mathcal{P}(i) = \{p \in \mathcal{B} \mid c_p = c_i, p \neq i\}$ is the set of samples in the same class as $i$, excluding itself. The total encoder training loss is defined as 
$\mathcal{L}_{\mathrm{enc}} = \mathcal{L}_{\mathrm{CE}} + \lambda_\mathrm{con}\mathcal{L}_{\mathrm{con}}$,
where $\lambda_{\mathrm{con}}$ is a hyper-parameter weighting the contrastive loss. 

\textbf{Quality Prototype Learning.}
Subsequently, we train the quality prototypes with the IQA encoder frozen. We optimize two prototype vectors, $\mathbf{p}_0 \in \mathbb{R}^{d}$ and $\mathbf{p}_1 \in \mathbb{R}^{d}$, corresponding to undistorted and distorted images, respectively. For each sample $(x_i, c_i)$, the extracted feature $\mathbf{z}_i$ is optimized to be closer to its class prototype and further from the opposing prototype. We utilize cosine similarity as the metric and apply a margin ranking loss with a constant margin $m$, defined as 
$\mathcal{L}_{\mathrm{proto}} = \frac{1}{|\mathcal{B}|}\sum_{i=1}^{|\mathcal{B}|} \max\left(0, \cos(\mathbf{z}_i, \mathbf{p}_{1-c_i}) - \cos(\mathbf{z}_i, \mathbf{p}_{c_i}) + m \right)$. 

\subsection{Weakly-Supervised Prototype FM Model Training}
Let $\mathbb{X}=\{x_i\}_{i=1}^{N_s}$ denote the source domain (undistorted DWI) and $\mathbb{Y}=\{y_j\}_{j=1}^{N_t}$ denote the target domain (distorted DWI), where $N_s$ and $N_t$ represent the number of samples in the source and target training sets, respectively. 
Since our ultimate clinical objective is distortion correction (forward direction, $\mathbb{Y} \rightarrow \mathbb{X}$), we formulate the distortion simulation as a reverse generation process ($\mathbb{X} \rightarrow \mathbb{Y}$).
The objective of FM is to learn a time-dependent velocity field $v_{\theta}(\cdot,t)$ that approximates the true conditional flow at any time $t \in [0,1]$.
During training, mini-batches of samples are drawn independently from $\mathbb{X}$ and $\mathbb{Y}$, and a continuous time step $t$ is sampled from the uniform distribution $\mathcal{U}(0,1)$. 
We employ a minibatch optimal transport (OT) plan, where source and target samples within each mini-batch are coupled to minimize the overall transport cost, as detailed in~\cite{tong2023improving}. Given an OT-coupled pair $(x_i, y_j)$ and a time step $t$, the intermediate state is defined via linear interpolation as $\hat{y}_{i,j}^t = (1 - t)x_i + t y_j$, with the ground-truth velocity field $u_{i,j} = y_j - x_i$.
The FM network takes $\hat{y}_{i,j}^t$ and $t$ as inputs to get the velocity field prediction $\hat{v}_{\theta}(\hat{y}_{i,j}^t, t)$.
The training objective minimizes the mean squared error between the predicted and ground-truth velocity fields as 
$\mathcal{L}_{\mathrm{FM}} = \mathbb{E}_{x_i,y_j,t} \left[\|\hat{v}_{\theta}(\hat{y}_{i,j}^t,t)-u_{i,j}\|_2^2\right]$.

In addition to learning the velocity field in the image space, we introduce a weak supervision strategy that explicitly guides the generated features toward the distorted feature manifold in the feature space.
The FM network generates a trajectory $\{x_i^{k}\}_{k=0}^{K}$ starting from the initial state $x_i^{0}$ (where $x_i^{0}=x_i$) with discrete time steps $t_k = k/K$.
We define a threshold step $k_0$. For early generation steps ($k < k_0$), we apply no quality constraints, as the early generated images are semantically unstable and noisy. 
For later steps ($k \geq k_0$), the generated image $x_i^{k}$ is passed through the IQA encoder at each step to extract its feature $\mathbf{z}_i^{k}$. We calculate a margin ranking loss between this feature and the quality prototypes to formulate the guidance loss. Crucially, we introduce a time-dependent weight $w_k$ to allocate stronger supervision as the generated image progressively reveals structural details, preventing the loss from misguiding the trajectory during the early, noisy stages. The prototype guidance loss is defined as:
\begin{equation}
\label{eq:fm_proto_loss}
\mathcal{L}_{\mathrm{PG}} = \sum_{k=k_0}^{K} w_k \cdot \max\left(0, \cos(\mathbf{z}_i^{k}, \mathbf{p}_0) - \cos(\mathbf{z}_i^{k}, \mathbf{p}_1) + m \right),
\end{equation}
where $w_k = [(k - k_0)/(K - k_0)]^2$ is the quadratic time-dependent weight and $m$ is the constant margin.
To maintain memory efficiency while optimizing this continuous trajectory, we employ gradient accumulation, updating the model parameters at intervals of $n_{\Delta k}$ steps.
The total training objective for the FM network is defined as $\mathcal{L}_{\mathrm{total}} = \mathcal{L}_{\mathrm{FM}} + \lambda_{\mathrm{PG}} \mathcal{L}_{\mathrm{PG}}$,
where $\lambda_{\mathrm{PG}}$ is a hyper-parameter that balances the prototype guidance loss.

\subsection{Inference and Second IQT Model Training.}
During inference, the FM network generates a distorted image iteratively starting from an undistorted source image $x_i$. For each discrete time step $k=1,\ldots,K$ (corresponding to $t_k = k/K$), 
the state $x_i^{k}$ is updated from the previous state $x_i^{k-1}$ using the predicted velocity field $\hat{v}_{\theta}(x_i^{{k-1}}, t_{k-1})$, as $x_i^{k} = x_i^{{k-1}} + \hat{v}_{\theta}(x_i^{{k-1}}, t_{k-1})\Delta t$, where $\Delta t = 1/K$. 
By applying this generation process to the undistorted dataset, we synthesize a paired dataset of undistorted and generated distorted images, denoted as $\{(x_i, \hat{y}_i)\}_{i=1}^{N_s}$, where $\hat{y}_i = x_i^{K}$ represents the final generated state. Utilizing these synthetic pairs, we train a supervised second IQT network denoted as $G_{\psi}$ to map the distorted DWI $\hat{y}_i$ back to the undistorted one $x_i$. The network is trained to minimize the reconstruction loss between the predicted undistorted image and the ground truth as $\mathcal{L}_{\mathrm{corr}} = \mathbb{E}_{x_i, \hat{y}_i} [ \| x_i - G_{\psi}(\hat{y}_i) \|_1 ]$, where $\|\cdot\|_1$ denotes the $L_1$ norm to preserve high-frequency textural details, which are critical for downstream clinical evaluations.

\section{Experiments and Results}

\textbf{Implementation Details.}
The FM and the second IQT models both employ 3D U-Net backbones~\cite{cciccek20163d} featuring a three-level symmetric encoder-decoder structure. Specifically, we set the channel multipliers to (1, 2, 4) with two residual blocks per level and integrate a self-attention module at the second level. The IQA encoder is implemented using a 3D ResNet-18 architecture~\cite{hara2018can}, which yields a 512-dimensional feature vector via global average pooling. For encoder pre-training, the temperature parameter $\tau$ and contrastive loss weight $\lambda_{\mathrm{con}}$ are set to $0.1$ and $0.5$, respectively. In the quality prototype learning and FM training stages, the margin $m$ is set to $0.2$. FM training utilizes $K=20$ total inference steps, with a threshold step $k_0=10$, an interval $n_{\Delta k}=3$, and a guidance loss weight $\lambda_{\mathrm{PG}}=1.0$. All models are optimized for 100 epochs using Adam with a learning rate of $1\times10^{-4}$ and a batch size of 4. Experiments are conducted on an NVIDIA Quadro GV100 GPU with 24 GB memory.

\textbf{Datasets.}
A private dataset~\cite{yan2024combiner}, referred to as 'Dataset 1', consisting of $1027$ prostate DWI scans with multiple high $b$-values ($\geq 1400$ s/mm$^2$) from $851$ patients, is used for training and internal testing. These cases are categorized into undistorted ($790$ scans) and distorted ($237$ scans) based on geometric consistency between T2WI and the imaging-coordinate-aligned DWI. Dataset 1 is randomly partitioned into training and test sets with an $8:2$ ratio at the patient level.
The PRIME dataset~\cite{asif2023comparing} is used for external evaluation, comprising $483$ cases, including $28$ labeled as distorted by expert radiologists. 
Additionally, it includes expert-labeled PI-RADS~\cite{turkbey2019prostate} ($227$ $\geq 4$, $256$ $< 4$) and Gleason scores~\cite{epstein2010update} ($209$ $\geq 3+4$, $274$ $< 3+4$) for downstream clinical validation.
All volumes are preprocessed using SimpleITK~\cite{lowekamp2013design} python library, resampled to a spatial resolution of $[1.0, 1.0, 1.5]$ mm, followed by center-cropping to $[128, 128, 32]$. Voxel intensities are normalized to the range $[-1, 1]$ via min-max scaling.

\textbf{Downstream Evaluation Protocol.}
We utilize the ProFound foundation model\footnote{\url{https://github.com/pipiwang/ProFound}} to evaluate the clinical value of generated images through downstream tasks. 
The model takes three modalities as input: high $b$-value DWI, axial T2WI, and the apparent diffusion coefficient (ADC) map. The ProFound model is fine-tuned on $386$ undistorted PRIME cases and tested on the remaining $60$ scans, which include $30$ randomly selected samples per class for each task.
During fine-tuning, affine transformations are applied jointly to all three modalities to preserve spatial alignment, including random rotation (up to $15^\circ$) and translation (up to 0.15 mm).
For comparison and ablation study evaluation, we replace the original undistorted DWI with generated distorted DWI to measure the performance degradation. 
For correction evaluation, we use $37$ real-world distorted volumes from the PRIME dataset and replace them with generated undistorted DWI.
Performance is quantified using accuracy (ACC) and area under the curve (AUC).
We follow the open-source ProFound fine-tuning code and use the same hyperparameter settings.

\begin{table*}[t]
\centering
\caption{Quantitative evaluation on the PRIME dataset. Comparison and Ablation Study: Performance of the downstream classifier when the original undistorted DWI is replaced with generated distorted DWI. Lower ACC and AUC indicate more effective distortion simulation. Distortion Correction: Evaluation of diagnostic correction on real distorted images. For unpaired baselines (C-GAN, U-DDPM, OT-FM) and Ours (FG), correction is performed via forward generation (distorted-to-undistorted). Ours (UNet) refers to a supervised U-Net trained on the synthetic paired data produced by our distortion synthesis FM model. Higher ACC and AUC indicate superior correction. Ref. Un and Ref. Dis denote reference performance on original undistorted and distorted test images, respectively.
All metrics are reported as mean $\pm$ standard deviation over five runs.
For diffusion and FM-based methods, results with total inference steps of $20$.}
\label{biao1}
\resizebox{0.9\textwidth}{!}{%
\begin{tabular}{l l c c c c}
\toprule
&  & \multicolumn{2}{c}{\makecell{PI-RADS Score}} & \multicolumn{2}{c}{\makecell{Gleason Score}} \\
\cmidrule(lr){3-4} \cmidrule(lr){5-6}
\makecell{Category}&\makecell{Method} & \makecell{ACC ($\%$)} & \makecell{AUC ($\%$)} & \makecell{ACC ($\%$)} & \makecell{AUC ($\%$)} \\
\midrule
\multirow{5}{*}{\makecell{Comparison \\ Study}} & \makecell{Ref. Un} & \makecell{59.33 $\pm$ 3.83} & \makecell{67.22 $\pm$ 3.98} & \makecell{57.33 $\pm$ 2.98} & \makecell{62.89 $\pm$ 2.11} \\
& \makecell{C-GAN} & \makecell{50.33 $\pm$ 4.41} & \makecell{56.02 $\pm$ 4.42} & \makecell{54.67 $\pm$ 4.14} & \makecell{47.24 $\pm$ 4.12} \\
& \makecell{U-DDPM} & \makecell{60.00 $\pm$ 4.08} & \makecell{57.11 $\pm$ 4.25} & \makecell{52.67 $\pm$ 2.88} & \makecell{52.91 $\pm$ 3.49} \\
& \makecell{OT-FM} & \makecell{57.67 $\pm$ 4.14} & \makecell{61.44 $\pm$ 5.77} & \makecell{56.33 $\pm$ 4.82} & \makecell{54.67 $\pm$ 5.56} \\
& \makecell{Ours} & \makecell{49.67 $\pm$ 2.11} & \makecell{53.78 $\pm$ 3.45} & \makecell{46.67 $\pm$ 3.54} & \makecell{45.82 $\pm$ 3.21} \\
\midrule
\multirow{3}{*}{\makecell{Ablation\\ Study}} & \makecell{w.o. PG} & \makecell{57.67 $\pm$ 4.14} & \makecell{61.44 $\pm$ 5.77} & \makecell{56.33 $\pm$ 4.82} & \makecell{54.67 $\pm$ 5.56} \\
& \makecell{PG $\rightarrow$ CE} & \makecell{54.33 $\pm$ 2.36} & \makecell{57.67 $\pm$ 3.12} & \makecell{58.33 $\pm$ 4.63} & \makecell{59.33 $\pm$ 4.23} \\
& \makecell{Ours} & \makecell{49.67 $\pm$ 2.11} & \makecell{53.78 $\pm$ 3.45} & \makecell{46.67 $\pm$ 3.54} & \makecell{45.82 $\pm$ 3.21} \\
\midrule
\multirow{6}{*}{\makecell{Distortion\\ Correction}}
& \makecell{Ref. Dis} & \makecell{42.86 $\pm$ 6.68} & \makecell{46.78 $\pm$ 8.76}  & \makecell{48.57 $\pm$ 9.65} & \makecell{48.65 $\pm$ 7.71} \\
& \makecell{C-GAN} & \makecell{48.57 $\pm$ 6.96} & \makecell{49.56 $\pm$ 6.44} & \makecell{55.00 $\pm$ 6.49} & \makecell{50.41 $\pm$ 9.63} \\
& \makecell{U-DDPM} & \makecell{41.43$\pm$ 8.22} & \makecell{42.78 $\pm$ 8.35} & \makecell{48.57 $\pm$ 8.22} & \makecell{50.53 $\pm$ 9.50} \\
& \makecell{OT-FM} & \makecell{45.71 $\pm$ 8.53} & \makecell{48.11 $\pm$ 8.97} & \makecell{43.57 $\pm$ 8.22} & \makecell{51.93 $\pm$ 9.50} \\
& \makecell{Ours (FG)} & \makecell{44.29 $\pm$ 5.98} & \makecell{47.33 $\pm$ 6.13} & \makecell{46.43 $\pm$ 7.99} & \makecell{46.08 $\pm$ 4.97} \\
& \makecell{Ours (UNet)} & \makecell{54.29 $\pm$ 4.66} & \makecell{53.98 $\pm$ 3.02} & \makecell{56.43 $\pm$ 5.42} & \makecell{51.11 $\pm$ 4.85} \\
\bottomrule
\end{tabular}%
}
\end{table*}

\begin{figure*}[t] \centering \includegraphics[width=0.95\textwidth]{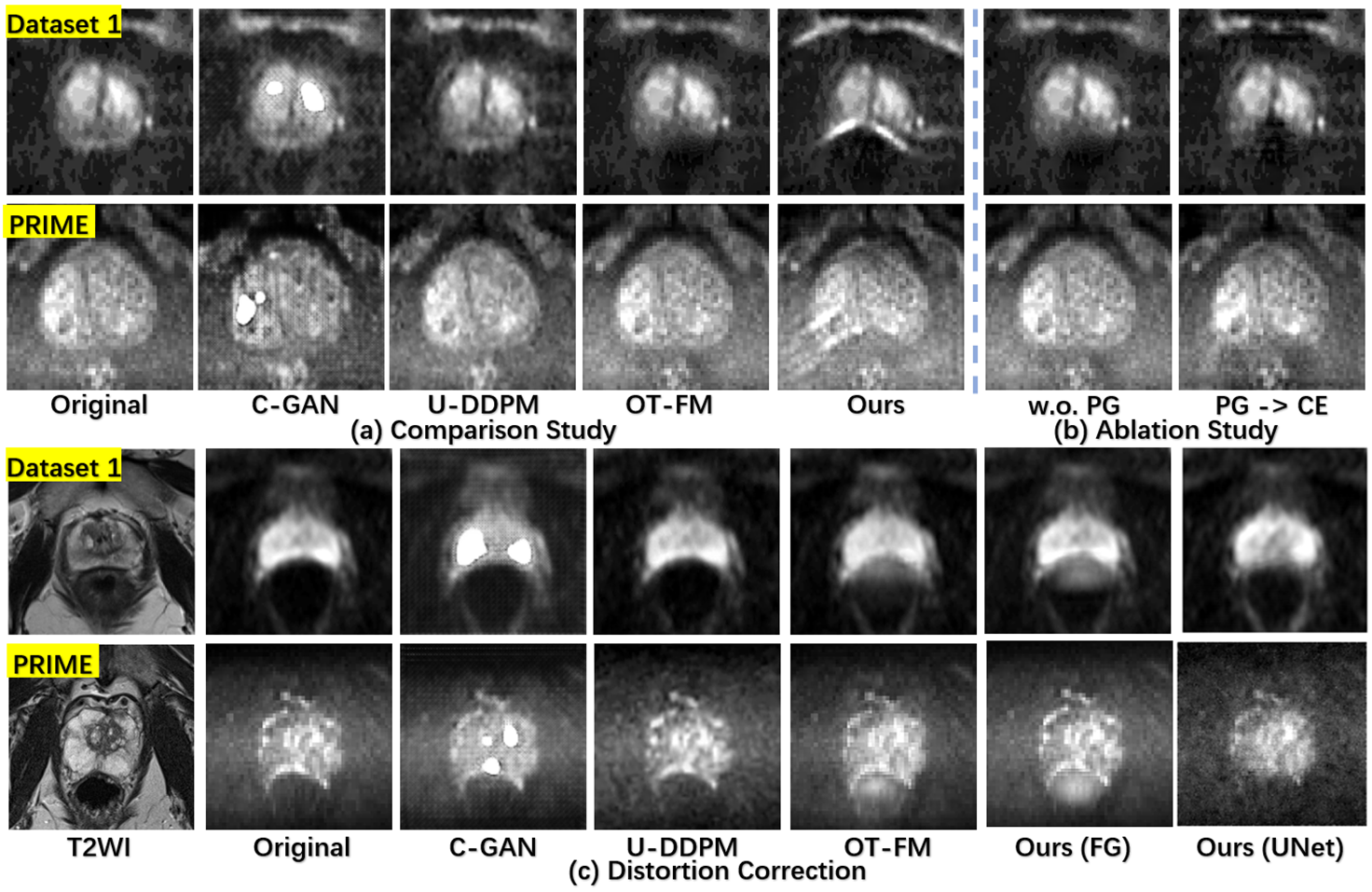} \caption{Qualitative evaluation on the in-distribution and external test datasets. (a) Visual comparison of distortion generation with baseline methods. (b) Ablation study on the weakly-supervised PG mechanism. (c) Visualization of baseline methods for direct forward distortion correction and the U-Net model trained on synthetic paired data (Ours (UNet)). } \label{tu2} \end{figure*}

\textbf{Comparison Study}
We compare the proposed weakly-supervised prototype FM method with CycleGAN (C-GAN)~\cite{sandfort2019data}, UNIT-DDPM (U-DDPM)~\cite{sasaki2021unit} using the SDEdit strategy~\cite{meng2021sdedit}, and optimal transport flow matching (OT-FM)~\cite{yazdani2025flow} on the test sets. All models share the same 3D U-Net backbone. U-DDPM, OT-FM, and our method use $20$ inference steps. Results are summarized in Table~\ref{biao1}, and compared in Fig.~\ref{tu2} (a). 
As shown in Table~\ref{biao1}, CycleGAN achieves relatively low classification scores, and primarily performs intensity and style transfer rather than learning geometric deformation as observed in Fig.~\ref{tu2} (a).
U-DDPM with SDEdit fails to simulate geometric distortions. It maintains PI-RADS performance by preserving anatomical structures, but significantly degrades Gleason scoring. This is perhaps due to the denoising process smoothing out fine textures and intensities essential for grading, reflecting textural degradation rather than realistic distortion.
For FM-based methods, the OT-FM yields higher ACC and AUC scores.
Visually, OT-FM primarily introduces blurring effects through pixel-wise averaging and fails to simulate complex structural degradations.
Without a defined target in the feature space, the standard flow trajectory tends to converge to a conservative mean representation of the target domain, failing to reach the regions of high distortion intensity within the manifold.
Our method is able to push velocity field toward the core distortion features, producing PI-RADS-and-Gleason-impacting boundary deformation.

\textbf{Ablation Study.}
We evaluate the weakly supervised prototype guidance (PG) mechanism by comparing our model with: (1) a version without the PG constraint (w.o. PG), which is equivalent to OT-FM, and a version where the PG loss is replaced by a standard binary cross-entropy (BCE) classification loss (PG $\rightarrow$ CE). The BCE loss is computed using the outputs of the pre-trained classification head in Section 2.1 against the label for distorted images ($c=1$).
As shown in Table~\ref{biao1} and Fig.~\ref{tu2} (b), both removing and substituting the PG constraint lead to a reduction in distortion intensity.
The CE approach is less effective than PG because the BCE loss is primarily optimized by pushing features across the decision boundary. Once the sample is classified as distorted, the gradient signal diminishes, often leaving the generated image near the boundary with insufficient geometric deformation. In contrast, by maximizing the cosine similarity to this cluster center, the PG constraint provides a continuous and stable directional force that pulls the generation trajectory deeper into the distorted distribution, resulting in more clinically realistic distortion.

\textbf{Supervised Distortion Correction via Synthetic Paired Data.}
As shown in Table~\ref{biao1}, direct forward generation (FG) ($\mathbb{Y} \rightarrow \mathbb{X}$) using the proposed approach (Ours (FG)) and other unpaired baselines yield poor downstream performance. Qualitatively, in Fig.~\ref{tu2}(c), the Ours (FG) merely fills geometric voids with unrealistic intensities. 
On the other hand, as presented in Table~\ref{biao1}, the U-Net trained on our synthetic data produced via reverse generation (Ours (UNet)) outperforms all baseline methods attempting direct correction on PI-RADS score task. 
However, we observe that the U-Net outputs tend to over-smooth high-frequency textures, resulting in blurrier images compared to the original undistorted references. This loss of fine-grained detail limits the model's performance on the Gleason score task. 
We expect further development using generative modeling for this step may warrant improvement.

\section{Conclusion}
We proposed a novel, practical and effective two-stage IQT framework for unpaired DWI distortion correction. Experiments confirm that the generated images are clinically realistic and downstream impactful. Furthermore, using these synthetic paired samples, a simple supervised distortion correction model improves image quality in real-world scenarios. 
Our future work will focus on employing more powerful generative models (such as FM) for the supervised correction stage to enhance textural fidelity. Furthermore, we plan to incorporate corresponding T2WI as structural priors to provide stronger anatomical constraints during the correction process.

\subsubsection{Disclosure of Interests.} The authors have no competing interests to declare that are relevant to the content of this article.
\bibliographystyle{splncs04}
\bibliography{bib.bib}
\end{document}